\documentclass{article}
\usepackage{spconf,amsmath,graphicx}
\usepackage{subfig}
\usepackage{bbold}

\title{Locality Preserving Multiview Graph Hashing for Large Scale Remote Sensing Image Search}
\name{Wenyun Li,  Guo Zhong, Xingyu Lu, Chi-Man Pun\sthanks{Corresponding author, email: cmpun@umac.mo. This work was supported in part by the University of Macau under Grant MYRG2022-00190-FST and in part by the Science and Technology Development Fund, Macau SAR,
under Grant 0034/2019/AMJ, Grant 0087/2020/A2, and Grant 0049/2021/A.}}
\address{Department of Computer and Information Science, University of Macau, Macau, China}
\begin{document}
%
\maketitle
\begin{abstract}
Hashing is very popular for remote sensing image search.  This article proposes a multiview hashing with learnable parameters to retrieve the queried images for a large-scale remote sensing dataset. Existing methods always neglect that real-world remote sensing data lies on a low- dimensional manifold embedded in high-dimensional ambient space.  Unlike previous methods, this article proposes to learn the consensus compact codes in a view-specific low-dimensional subspace. Furthermore, we have added a hyperparameter learnable module to avoid complex parameter tuning. In order to prove the effectiveness of our method, we carried out experiments on three widely used remote sensing data sets and compared them with seven state-of-the-art methods. Extensive experiments show that the proposed method can achieve competitive results compared to the other method.
\end{abstract}
\section{Introduction}
\label{sec:intro}
With the development of remote sensing (RS) technology, large-scale RS image data has attracted extensive attention, but the processing, fusion, and mining of large-scale images is a huge problem. As a critical problem of RS, remote sensing image search (RSIS) plays a vital role in the meteorological analysis\cite{dell2001query}, geographic analysis\cite{yu2014full} and earth observation image process\cite{zhu2019matching}. RSIS aims to find the image that most closely resembles the query image. To efficiently search the queried images and boost the retrieval performance, hashing for RSIS has attracted extensive attention in the community\cite{slaney2008locality}. The basic idea of hashing for RSIS is to project the high-dimensional multiview data into compact binary bits\cite{slaney2008locality}. Thanks to the bit-wise similarity measurement, hashing methods are very efficient in storage and computation. The existing remote sensing image search (RSIS) methods can be broadly divided into two categories. The first category focuses on how to extract practical features from RS images. The feature representations in early works\cite{bretschneider2002retrieval}  are often constructed by mining spectral, texture, or shape cues from RS images. Naturally, a lot of hand-crafted features which represent low-level vision information are used for RSIS, such as wavelet and Gabor filters\cite{daugman1988complete}. The second category focuses on indexing the feature in feature extractors of RS images. When the scale of the remote sensing image dataset becomes very large, feature indexing is adopted for handling massive RS data. It is not necessary when the RS image dataset is with a relatively small volume.

Most remote sensing multiview hashing methods can be classified into supervised and unsupervised categories. Specifically, supervised approaches\cite{lukavc2015gpu} typically use labeled semantic information to learn hash codes from multiview data, achieving promising performance. However, extensive data with label information is not easy to obtain. Unlike the supervised approaches\cite{slaney2008locality}, the unsupervised hash approaches do not require intensive data annotation, which is more attractive in practice. Compared to single-view hashing of RS images, multiview hashing methods can utilize multiple features in RS images to obtain higher performance. For example, representations like sift and gist features can indicate different levels of information on RS images.

In remote sensing, real-world data is mapped from a low-dimensional manifold to a high-dimensional space manifold\cite{he2003locality}. In order to depict such a manifold, the neighborhood relationship between data points is crucial. However, existing methods like\cite{xu2013harmonious}  cannot preserve the neighborhood relationships and measure the quantization loss end-to-end. Moreover, we employ a learnable parameter module to avoid additional parameters to balance the different values.  In order to solve the above issues, we propose an unsupervised multiview hashing method called Locality Preserving Multiview Graph Hashing (LPMGH). In comparison with previous work, our main contributions can be summarized as follows:
\begin{enumerate}
    \item We propose LPMGH, which simultaneously optimizes the average projection loss and the quantization loss. To our knowledge, this is one of the first efforts to learn multiview hash codes through joint optimization of projection and quantization stages to maintain the local structure of data sets in the remote sensing community.
    \item We develop an efficient parameter learnable hashing method, which avoids complex hyperparameter tuning. All balance parameters are obtained during training. LPMGH can balance the influence of the weights of different views on RS data.
    \item We have conducted extensive experiments on three widely used remote-sensing image datasets. The results show that the proposed LPMGH is superior to the state-of-the-art multi view hash method in RSIS.
\end{enumerate}

\section{Proposed Method}
\label{sec:format}
\begin{figure}[h]
\centering
\includegraphics[width=8.5cm]{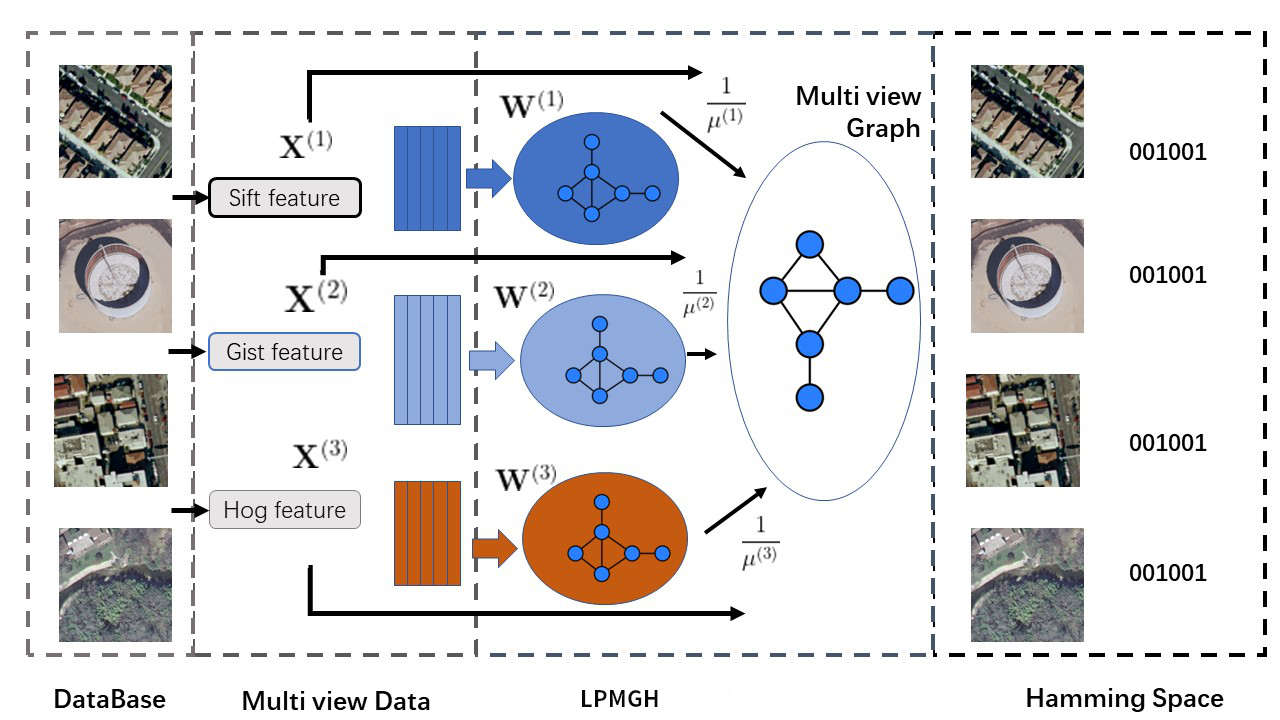} 
\caption{ The proposed Locality Preserving Multiview Graph Hashing (LPMGH) framework.} 
\label{framework}
\end{figure}
Fig.\ref{framework} shows the architecture of our proposed network. Given the visual features $\left \{ \mathbf{X}^{\left ( m \right )} = [\mathbf{x}_{1}^{\left ( m \right )} ,\ldots ,\mathbf{x}_{n}^{\left ( m \right )}] \right \}$ for $m$ views, the proposed method can generate the hashcode $\mathbf{B}$.
\subsection{Formulation}
In LPMGH, we utilize spectral hashing\cite{weiss2008spectral} for similarity preservation. For data matrix $\mathbf{X} \in \mathbb{R} ^{N \times d}$ the hashing formulation as follows:
\begin{equation}
\label{eqn_1}
\begin{array}{c}
min \sum _{i,j}\mathbf{A}_{i,j}\left \| \mathbf{b}_i - \mathbf{b}_j\right \|^{2} \\
\\
s.t. \sum _{i}\mathbf{b}_i=0,\qquad \frac{1}{n}\sum _{i}\mathbf{b}_i \mathbf{b}_i^{T}=\mathbf{I}_{r},\quad \mathbf{b}_i \in \left \{ -1,1 \right \}^{r}
 \end{array}
\end{equation}
\begin{equation}
\label{eqn_2}
\begin{array}{c}
\underset{\mathbf{B}}{argmin} Tr\left ( \mathbf{B}^{T}\left ( \mathbf{D}-\mathbf{A} \right )\mathbf{B} \right ) \\
s.t. \mathbf{B}^{T}\mathbf{B}=n\mathbf{I}_r,\quad \mathbf{B}^{T}\mathbf{1}_{n\times 1}=0,\quad  \mathbf{B}\in \left \{ -1,1 \right \}^{n \times r}
 \end{array}
\end{equation}
It is worth noting that $\mathbf{A}$ is a low-rank PSD matrix with rank at most $P$, so we can have a low-rank approximation as follows
\begin{equation}
\label{eqn_3}
\mathbf{L}=\mathbf{I}-\mathbf{Z}\Lambda ^{-1}\mathbf{Z}^{T}
\end{equation}
$\mathbf{A}$ is always sparse, so we have $\mathbf{A}=\mathbf{Z}\mathbf{\Lambda }^{-1}\mathbf{Z}^{T}$ with $\mathbf{\Lambda }=diag\left ( \mathbf{Z}^{T}\mathbf{1} \right )$, and $\mathbf{D}\approx \mathbf{I}$. In single view case, we set the relaxed real values $\mathbf{B'}$ to support later binary computation. Since remote sensing data is located on a low dimensional manifold, the neighborhood structure of the manifold should be preserved to capture meaningful neighborhood through view specific hash. Here we adopt a matrix to project and maintain the local structure simultaneously.  So we assume as follows:
\begin{equation}
\label{eqn_5}
\begin{array}{c}
\mathbf{B'}=\mathbf{X}\mathbf{W}
\\
s.t. \mathbf{W}^{T}\mathbf{W}=\boldsymbol{I}_r
 \end{array}
\end{equation}
where $\mathbf{W}$ is an orthogonal matrix. The optimization problem will convert into the following problem:
\begin{equation}
\label{eqn_6}
\begin{array}{c}
\underset{\mathbf{W}}{min}-Tr\left ( (\mathbf{XW})^{T}\mathbf{A}(\mathbf{XW}) \right )
\\
s.t. (\mathbf{XW})^{T}(\mathbf{XW})=n\mathbf{I}_{r}, \\
(\mathbf{XW})^{T}\mathbf{1}_{n\times 1}=\mathbf{0}.
 \end{array}
\end{equation}

We denote $\mathbf{S}=\mathbf{X}^{T}\mathbf{A}\mathbf{X}$. Note that Eq.\ref{eqn_6} is the single view case, now we can extend it to multi-view hashing as follows:

\begin{equation}
\label{eqn_8}
\begin{array}{c}
\underset{\mathbf{W} ^{(m)}}{min}-\sum _{m=1}^{M}Tr\left ( \mathbf{W}^{(m)T}\mathbf{S}^{(m)}\mathbf{W}^{(m)} \right )
\\
s.t. \mathbf{W} ^{(m)T}\mathbf{W} ^{(m)}=\mathbf{I}_{r}.
 \end{array}
\end{equation}

Inspired by \cite{zheng2020efficient}, we adopt a self-learnable weight parameter $\mu ^{\left ( m \right )}$ for the $m$-th view. So Eq.\ref{eqn_5} extend to multi-view cases in binary condition and combines with Eq.\ref{eqn_5}, so the overall object function as follows:
\begin{equation}
\label{eqn_10}
\begin{array}{c}
\underset{\mathbf{W} ^{(m)},\mathbf{B},\mu ^{\left ( m \right )}}{min} \sum _{m=1}^{M}-Tr\left ( \mathbf{W}^{(m)T}\mathbf{S}^{(m)}\mathbf{W}^{(m)} \right )+ \\ \frac{1}{\mu ^{\left ( m \right )}}\left \| \mathbf{B}-\mathbf{X}^{(m)}\mathbf{W} ^{(m)} \right \|_{F}^{2}
\\
s.t. \mathbf{W} ^{(m)T}\mathbf{W} ^{(m)}=\mathbf{I}_{r}, \\
\mathbf{B}\in \left \{ -1,1 \right \}^{n \times r}.
 \end{array}
\end{equation}
\subsection{Optimization}
\label{op}
Since the object function Eq.\ref{eqn_10} is in discrete and orthogonal condition, it is hard to optimize. We employ the iterative algorithm to optimize the variables as follows:
\begin{enumerate}
    \item \label{itm:first} \textit{Initialize}: To initialize $\mathbf{W} ^{(m)}$, we calculate the $r$ eigenvectors of the similarity matrix $\mathbf{S} ^{(m)}$ with the $r$ largest eigenvalues. For $\mathbf{B} $, we initialize it as a one-like matrix of size $n \times r$. 
    \item \label{itm:2} \textit{Update} $\mathbf{W} ^{(m)}$:
    First, we fixed the variables $\mathbf{B} $ , $\mu^{(m)} $ and update $\mathbf{W} ^{(m)}$.
    It is tough to solve this problem due to the orthogonality of $\mathbf{W} ^{(m)}$. This orthogonal problem can be efficiently solved based on the method developed in\cite{wen2013feasible}.

\item \textit{Update} $\mu^{(m)}$:Here, we fixed the variables $\mathbf{B} $ and $\mathbf{W} ^{(m)}$ and updated $\mu^{(m)} $. The Eq.\ref{eqn_10} is changing to the following:
\begin{equation}
\label{eqn_15}
\underset{\mu^{(m)} }{min} \sum _{m=1}^{M}\frac{1}{\mu ^{\left ( m \right )}}\left \| \mathbf{B}- \mathbf{X}^{\left ( m \right )}\mathbf{W}^{\left ( m \right )} \right \|_{F}^2
\end{equation}
For simplicity, by denoting $l_{(m)}=\left \| \mathbf{B}-\mathbf{X}^{\left ( m \right )}\mathbf{W}^{\left ( m \right )} \right \|_{F}^2$, Eq.\ref{eqn_15} becomes:
\begin{equation}
\label{eqn_16}
\underset{\mu^{(m)} }{min} \sum _{m=1}^{M}\frac{1}{\mu ^{\left ( m \right )}}l_{(m)}
\end{equation}
Thus, we can obtain the closed solution of $\mu^{(m)} $ based on the method of \cite{zheng2020efficient} as follows:
\begin{equation}
\label{eqn_17}
\mu ^{\left ( m \right )}=\frac{l_{(m)}}{\sum _{i=1}^{M}l_{(m)}}
\end{equation}
\item \textit{Update} $\mathbf{B} $:With other variables fixed, the optimization  problem about $\mathbf{B} $ is reformulated as
\begin{equation}
\label{eqn_18}
\underset{\mathbf{B} }{min} \sum _{m=1}^{M}\frac{1}{\mu ^{\left ( m \right )}}\left \| \mathbf{B}- \mathbf{X}^{\left ( m \right )}\mathbf{W}^{\left ( m \right )} \right \|_{F}^2
\end{equation}
Clearly, the closed-form solution of $\mathbf{B}$ is easy to get as:
\begin{equation}
\label{eqn_20}
\mathbf{B}=sgn\left ( \sum _{m=1}^{M}\frac{\mathbf{X}^{\left ( m \right )}\mathbf{W}^{\left ( m \right )}}{\mu ^{\left ( m \right )}} \right )
\end{equation}
\end{enumerate}
\section{Experiments}
\label{sec:pagestyle}
\subsection{Implementation Details}
\begin{table}[!t]
\scriptsize
\caption{Statistics of the three datasets\label{tab:table1}}
\centering
\begin{tabular}{r|ccccc}
\hline
Datasets & Dataset & Training & Query& Image size & Dim     \\ \hline \hline
UCM      & 2,100   & 1,680    & 420  &256$\times$256 & 512/500 \\ 
NWPU     & 31,500  & 25,200   & 6,300 &256$\times$256  & 228/150 \\ 
AID      & 10,000  & 8,000    & 2,000 &600$\times$600  & 384/300 \\ \hline
\end{tabular}
\end{table}
Here we take three common datasets of RS into our evaluation, which include UCM\cite{yang2010bag}, NWPU\cite{cheng2017remote} and AID\cite{xia2017aid} datasets. Gist and Sift features are used to distinguish in multiview cases. The detail of the three datasets in our experiments is shown in Table\ref{tab:table1}.
Experiments were conducted on Intel i7-10700K CPU and 32 GiB RAM in Matlab. In our experiment, we set the initial value of $\mu ^{(m)}$ to 0.5 in section.\ref{op}. This article adopts two commonly used metrics: mean average precision (MAP) and precision-recall curve to quantitatively evaluate the search performance of large-scale remote sensing images.

\subsection{Performance Evaluation}
\label{sec:typestyle}
\begin{figure}
\centering
\subfloat{\includegraphics[width= 1.6in]{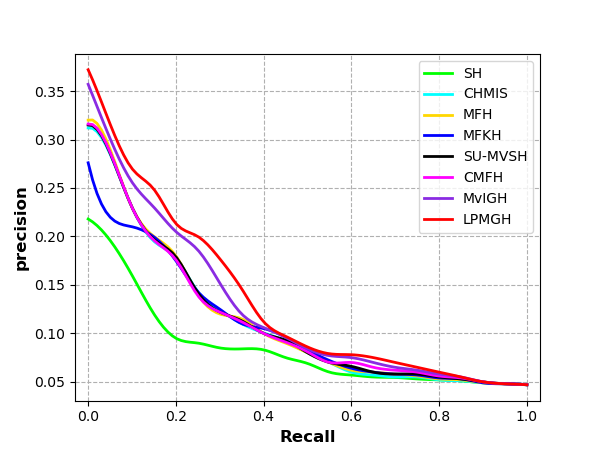}}
\subfloat{\includegraphics[width= 1.6in]{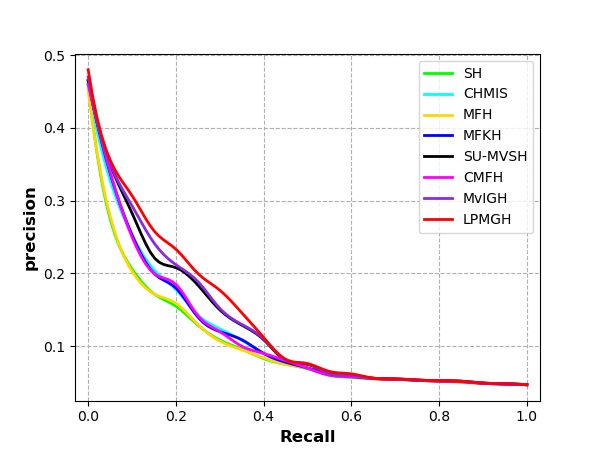}}
\\
\subfloat{\includegraphics[width= 1.6in]{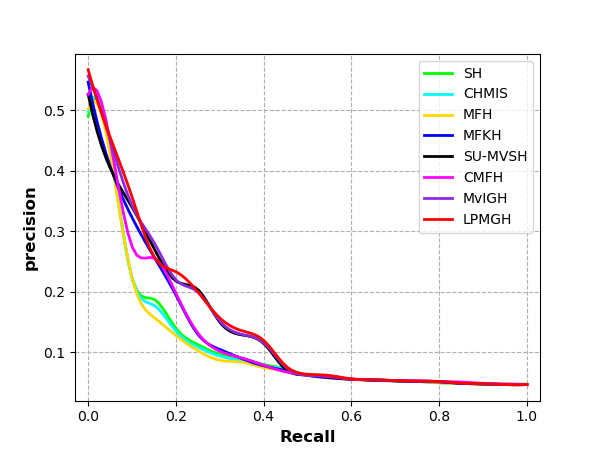}}
\subfloat{\includegraphics[width= 1.6in]{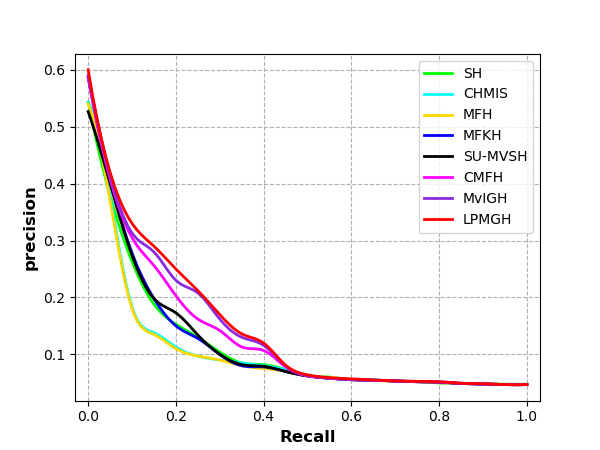}}
\caption{Precision-recall curves on UCM dataset of 16, 32, 64, 128 bits, respectively.}
\label{figs:UCM_data}
\end{figure}

To verify the effectiveness of our LPMGH, we compare it with a method SH\cite{weiss2008spectral} that relies on single-view hashing and six hashing methods based on multiview, including CHMIS\cite{zhang2011composite}, MFH\cite{song2011multiple}, MFKH\cite{liu2012compact}, SU-MVSH\cite{kim2012sequential}, CMFH\cite{6909664} and MvIGH\cite{9580682}. The mAP results of these methods on three benchmark datasets are shown in Tables.\ref{tab:UCM}-\ref{tab:AID}. The PR curves with different hash code lengths, including 16, 32, 64 and 128-bit on the UCM dataset are shown in Fig.\ref{figs:UCM_data}. The observations can be found from these results.

\begin{enumerate}
    \item The proposed LPMGH achieves the best performance on all benchmark datasets compared to other methods. LPMGH has the best mAP results in all the 12 cases from Tables \ref{tab:UCM}-\ref{tab:AID}. They show the effectiveness of LPMGH in retrieving the queried RS images.  In Fig.\ref{figs:UCM_data}, we find that the PR curves of LPMGH are above the others. These results clearly verify that LPMGH outperforms comparison methods in large-scale visual retrieval tasks.
    \item From Tables \ref{tab:UCM}-\ref{tab:AID}, we can see that multiview methods outperform the single-view methods in most cases. Furthermore, this verifies our previous assumption that increasing the number of views can help to achieve better performance. The effectiveness of the proposed LPMGH increases with the length of hash codes increasing due to more preserved information will take advantage of discrimination.
    \item From Table.\ref{tab:UCM}, we can find that the performance of LPMGH in the dataset UCM increases first when the code length increases from [16,32,64], then decreases when the code length is 128. One possible reason is the low variance of the latter bits in these methods, which degrades the quality of the overall hash code.
\end{enumerate}
\subsection{Convergence Analysis}
\begin{figure}[h]
\centering
\includegraphics[width=4.5cm]{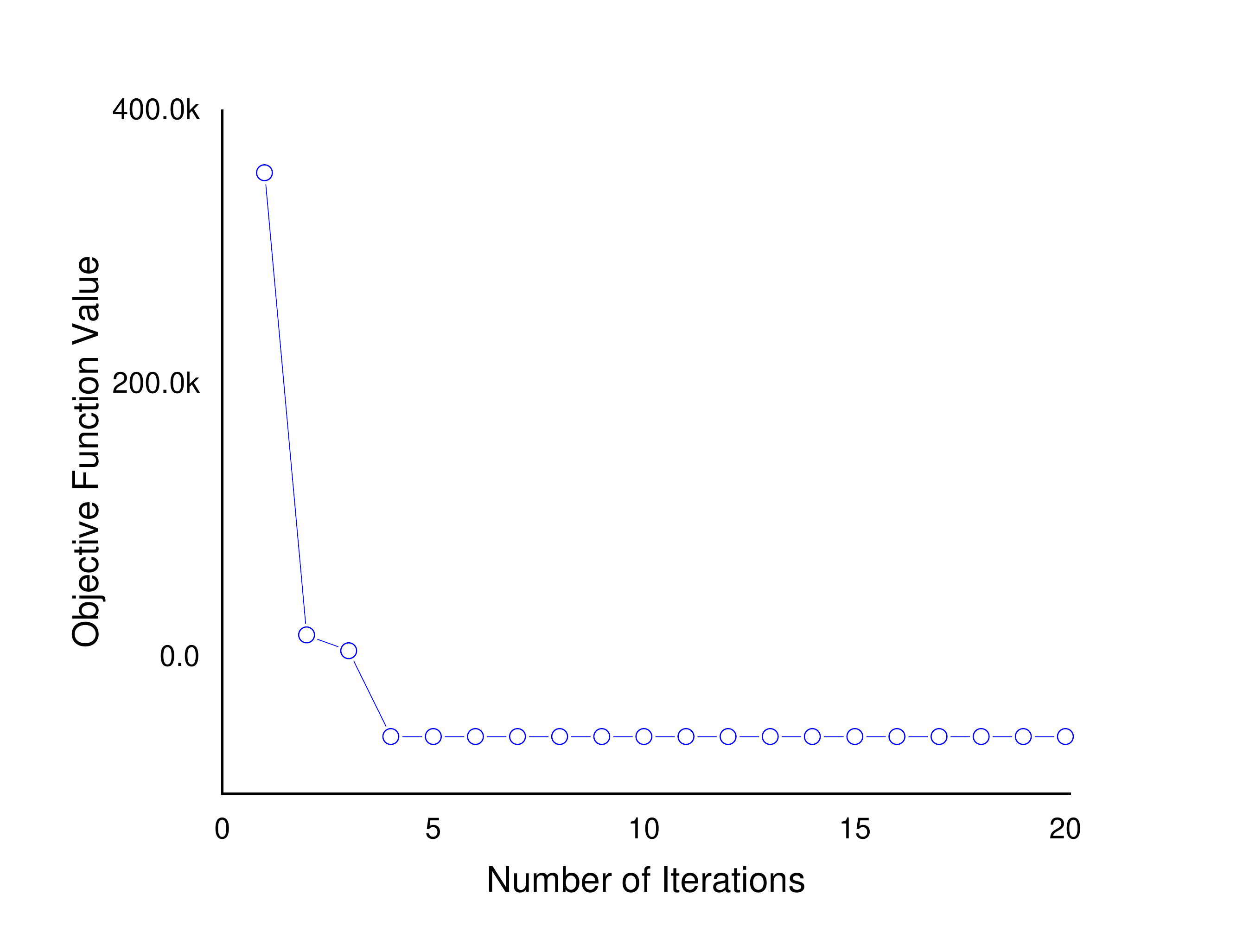} 
\caption{ Convergence analysis of the proposed LPMGH on the UCM dataset.} 
\label{iterations}
\end{figure}

We conduct an empirical analysis of the convergence of LPMGH. We implement an experiment on the UCM dataset. Fig.\ref{iterations} shows the convergence curves of LPMGH. From Fig.\ref{iterations}, we can clearly see that the convergence ability of LPMGH is outstanding in several iterations.
\subsection{Computation Complexity Analysis}
We evaluate the computational complexity of the proposed method on the NWPU dataset. We compared the training time of our method with other methods in 64-bit code, shown in Table\ref{tab:time}. We can observe that LPMGH is slower than SH, SU-MVSH, and CMFH and faster than other multiview hashing methods. The training time of LPMGH is spent chiefly on solving the orthogonal matrix.
\begin{table*}[]
\caption{Training Time on NWPU Dataset.\label{tab:time}}
\scriptsize
\centering
\begin{tabular}{c|cccccccc}
\hline
Method  & SH   & CHIMS   & MFH    & MFKH  & SU-MVSH & CMFH & MvIGH & LPMGH \\ \hline
Time(s) & 0.16 & 1090.59 & 590.66 & 20.77 & 1.42    & 4.59 & 29.04 & 14.31 \\ \hline
\end{tabular}
\end{table*}

\subsection{Influence of Hyper Parameters}
\begin{figure}[h]
\centering
\includegraphics[width=4cm]{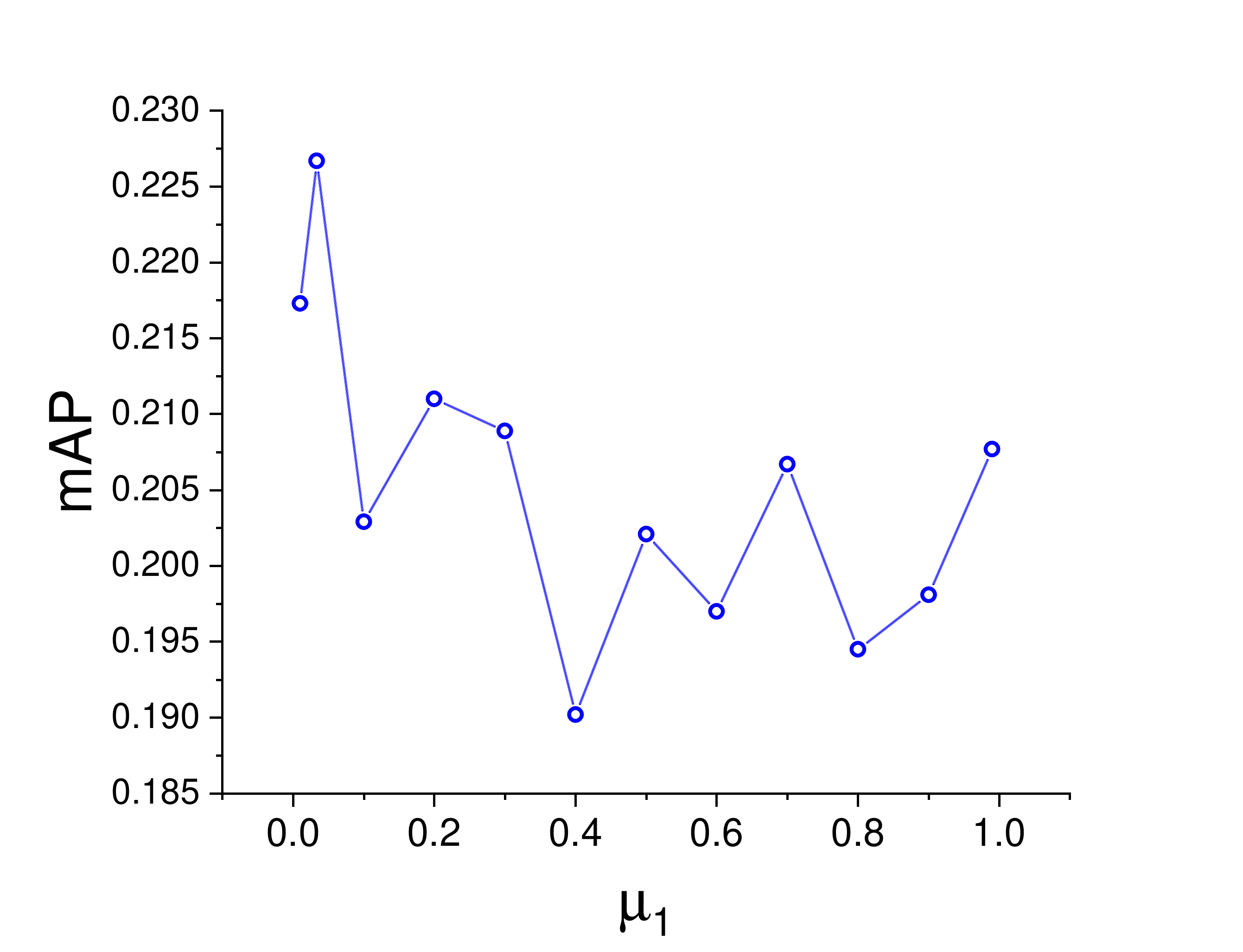} 
\caption{ Influence of $\mu_1$ in Locality Preserving Multiview Graph Hashing (LPMGH).} 
\label{mAP}
\end{figure}

Our method can learn the hyperparameter automatically in training and avoid parameter tuning. In order to analyze the influence of parameters which includes learning optimal value $\mu_1$ and other values. The range is from [0.01, 0.1, 0.2, 0.3, 0.5, 0.6, 0.7, 0.8, 0.9, 0.99]. Fig.\ref{mAP}  shows the MAP of different parameters in the 16-code-length-bit on UCM dataset. From the results, we can see the performance under different hyperparameters. Our method can easily obtain the optimal value in $\mu_1$ at 0.033, advancing in obtaining the optimal value compared to other fine-tuning-based methods.

\begin{table}[!t]
\scriptsize
\caption{MAP comparison of different length bits on the UCM dataset. The highest
score is shown in \textbf{boldface}.\label{tab:UCM}}
\centering
\begin{tabular}{r|cccc}
\hline
Method  & 16              & 32              & 64              & 128             \\ \hline
SH      & 0.1076          & 0.1417          & 0.1621          & 0.1569          \\ 
CHMIS   & 0.1612          & 0.1677          & 0.1499          & 0.1262          \\ 
MFH     & 0.1476          & 0.1427          & 0.1361          & 0.1251          \\ 
MFKH    & 0.1510          & 0.1618          & 0.1748          & 0.1482          \\ 
SU-MVSH & 0.1623          & 0.1901          & 0.2013          & 0.1711          \\ 
CMFH    & 0.1472          & 0.1639          & 0.1761          & 0.1881          \\ 
MvIGH   & 0.1832          & 0.1954          & 0.2110          & 0.2117          \\ \hline
LPMGH   & \textbf{0.2267} & \textbf{0.2341} & \textbf{0.2411} & \textbf{0.2169} \\ \hline
\end{tabular}
\end{table}
\begin{table}[!t]
\scriptsize
\caption{MAP comparison of different length bits on the NWPU Dataset. The highest
score is shown in \textbf{boldface}.\label{tab:NWPU}}
\centering
\begin{tabular}{r|cccc}
\hline
Method  & 16              & 32              & 64              & 128             \\ \hline
SH      & 0.0307          & 0.0309          & 0.0330          & 0.0373          \\ 
CHMIS   & 0.0444          & 0.0431          & 0.0378          & 0.0331          \\ 
MFH     & 0.0399          & 0.0399          & 0.0369          & 0.0330          \\ 
MFKH    & 0.0449          & 0.0497          & 0.0484          & 0.0453          \\ 
SU-MVSH & 0.0412          & 0.0496          & 0.0484          & 0.0453          \\ 
CMFH    & 0.0406          & 0.0428          & 0.0485          & 0.0513          \\ 
MvIGH   & 0.0459          & 0.0531          & 0.0543          & 0.0571          \\ \hline
LPMGH   & \textbf{0.0535} & \textbf{0.0574} & \textbf{0.0605} & \textbf{0.0625} \\ \hline
\end{tabular}
\end{table}
\begin{table}[!t]
\caption{MAP comparison of different length bits on the AID Dataset. The highest
score is shown in \textbf{boldface}.\label{tab:AID}}
\scriptsize
\centering
\begin{tabular}{r|cccc}
\hline
Method  & 16              & 32              & 64              & 128             \\ \hline
SH      & 0.0572          & 0.0556          & 0.0562          & 0.0660          \\ 
CHMIS   & 0.0832          & 0.0771          & 0.0662          & 0.0588          \\ 
MFH     & 0.0665          & 0.0637          & 0.0633          & 0.0589          \\ 
MFKH    & 0.0801          & 0.0879          & 0.0847          & 0.0793          \\ 
SU-MVSH & 0.0716          & 0.0893          & 0.1008          & 0.0959          \\ 
CMFH    & 0.0721          & 0.0810          & 0.0937          & 0.1004          \\ 
MvIGH   & 0.0867          & 0.0957          & 0.1017          & 0.1080          \\ \hline
LPMGH   & \textbf{0.0927} & \textbf{0.1024} & \textbf{0.1042} & \textbf{0.1094} \\ \hline
\end{tabular}
\end{table}

\section{Conclusion}
\label{sec:majhead}

This paper studies how to learn compact hash codes via anchor graph based hashing among multiview RS images. In this article, we propose a novel locality-preserving multiview graph hashing method(LPMGH), which  simultaneously  optimizes the average projection loss and quantization loss. In addition, we set all parameters learnable without introducing more hyper-parameters, which makes our method escape from complex parameter tuning and much more robust. Extensive experiments on three widely used remote sensing image datasets demonstrate that our method outperforms existing multi-view hashing methods on large-scale RSIS tasks.

Deep learning methods achieve remarkable performance due to their excellent ability to extract discerning features an automatic gradient optimization in computer vision. In the future, we will extend our method into deep neural networks. Furthermore, in reality, the RS data is not all available; it might generate noise. How to effectively mask the noisy information and utilize the limited RS images is challenging work.




\bibliographystyle{IEEEbib}
\bibliography{strings}

\end{document}